\documentclass{article}

\PassOptionsToPackage{numbers}{natbib}
\usepackage[preprint]{neurips_2026}
\usepackage{amsthm}
\usepackage{newunicodechar}
\usepackage{amsmath}     
\usepackage[most]{tcolorbox}                                      \usepackage{caption}     
\usepackage{amssymb}     
\usepackage{bm}          
\usepackage{graphicx}    

\newunicodechar{●}{\textbullet}
\usepackage{amsthm}                                                                                                                                              
    
\usepackage[utf8]{inputenc} 
\usepackage[T1]{fontenc}    
\usepackage{hyperref}       
\usepackage{url}            
\usepackage{booktabs}       
\usepackage{amsfonts}       
\usepackage{nicefrac}       
\usepackage{microtype}      
\usepackage{xcolor}         
\usepackage{amsmath,amssymb}
\usepackage{booktabs}
\usepackage{hyperref}
\usepackage{microtype}
\usepackage{xcolor}
\usepackage{pifont}
\usepackage{xcolor}

\usepackage{amsmath,amssymb}                                                                                                                           
\usepackage{booktabs}                                     
\usepackage{hyperref}                                                                                                                                  
\usepackage{microtype}                                        

\usepackage[T1]{fontenc}    
\usepackage{hyperref}       
\usepackage{url}            
\usepackage{amsfonts}       
\usepackage{nicefrac}       
\usepackage{microtype}      
\usepackage{xcolor}         
\usepackage{booktabs}
\usepackage{booktabs}
\usepackage{multirow}
\usepackage{graphicx}
\usepackage{colortbl}
\usepackage{xcolor}

\definecolor{yellowrow}{RGB}{255,245,200}
\definecolor{pinkrow}{RGB}{255,220,220}
\definecolor{bluerow}{RGB}{210,230,255}
\definecolor{greenrow}{RGB}{210,245,220}

\definecolor{grayrow}{RGB}{230,230,230}
\definecolor{orangrow}{RGB}{255,235,200}
\definecolor{purplerow}{RGB}{225,215,255}
\definecolor{tealrow}{RGB}{200,240,235}

\usepackage{enumitem}  
\title{SteerSeg: Attention Steering for Reasoning Video Segmentation}

\author{
  Ali Cheraghian$^{1,*}$ \quad
  Hamidreza Dastmalchi$^{2,*}$ \quad
  Abdelwahed Khamis$^{3}$ \quad
  Morteza Saberi$^{4}$ \quad \\
  \textbf{Aijun An}$^{2}$ \quad
  \textbf{Lars Petersson}$^{3}$ \\[0.4ex]
  $^{1}$Macquarie University \quad
  $^{2}$York University \quad
  $^{3}$CSIRO Data61 \quad
  $^{4}$UTS \\[0.3ex]
  \small
  \texttt{ali.cheraghian@mq.edu.au} \quad
  \texttt{hrd@yorku.ca} \quad
  \texttt{abdelwahed.khamis@data61.csiro.au} \\
  \texttt{morteza.saberi@uts.edu.au} \quad
  \texttt{aan@yorku.ca} \quad
  \texttt{lars.petersson@data61.csiro.au}
}

\begin{document}

\maketitle

\footnotetext{Equal contribution}
\begin{abstract}

Video reasoning segmentation requires localizing objects across video frames from natural language expressions, often involving spatial reasoning and implicit references. Recent approaches leverage frozen large vision-language models (LVLMs) by extracting attention maps and using them as spatial priors for segmentation, enabling training-free grounding. However, these attention maps are optimized for text generation rather than spatial localization, often resulting in diffuse and ambiguous grounding signals. 
In this work, we introduce \textbf{SteerSeg}, a lightweight framework that identifies attention misalignment as the key bottleneck in attention-based grounding and proposes to \emph{steer attention at its source} through input-level conditioning. SteerSeg combines learnable soft prompts with reasoning-guided Chain-of-Thought (CoT) prompting. The soft prompts reshape the attention distribution to produce more spatially concentrated maps, while CoT-derived attributes resolve ambiguity among similar objects by guiding attention toward the correct instance. The resulting attention maps are converted into point prompts across keyframes to guide a segmentation model, while candidate tracklets are ranked and selected using correlation-based scoring. 
Our approach freezes the LVLM and segmentation model parameters and learns only a small set of soft prompts, preserving the model’s pretrained reasoning capabilities while significantly improving grounding. Despite being trained only on Ref-YouTube-VOS, SteerSeg generalizes well across diverse benchmarks, significantly improving the spatial grounding capability of LVLMs. Project page: \href{https://steerseg.github.io/}{https://steerseg.github.io/}.

\end{abstract}


\section{Introduction}
\label{sec:intro}


Video reasoning segmentation aims to localize and segment objects in video frames given natural language queries requiring reasoning over spatial, temporal, and semantic relationships. While some queries can be resolved from direct visual cues, more complex expressions, such as \textit{``the object the child is reaching for''}, require understanding interactions and intent over time. These scenarios demand models that can both interpret language and reason about visual content consistently across frames, making video reasoning segmentation a challenging testbed for evaluating visual reasoning capabilities. Addressing this problem naturally calls for large vision-language models (LVLMs), which capture rich semantic and contextual relationships across modalities~\cite{khoreva2018video, seo2020urvos, ding2023mevis, yan2024visa, bai2024one}.

A common approach couples LVLMs with segmentation foundation models such as SAM~\cite{kirillov2023sam} in a cascaded framework, where the LVLM provides semantic guidance and the segmentation model predicts pixel-level masks. Methods such as LISA~\cite{lai2024lisa}, VISA~\cite{yan2024visa}, and VideoLISA~\cite{bai2024one}  train these cascaded pipelines end-to-end using large-scale annotated datasets. Although effective, such approaches require expensive supervised optimization, substantial labeled data, and re-training for each LVLM, even with parameter-efficient tuning methods such as LoRA~\cite{hu2022lora}.

\begin{figure}
    \centering
    \includegraphics[width=1\linewidth]{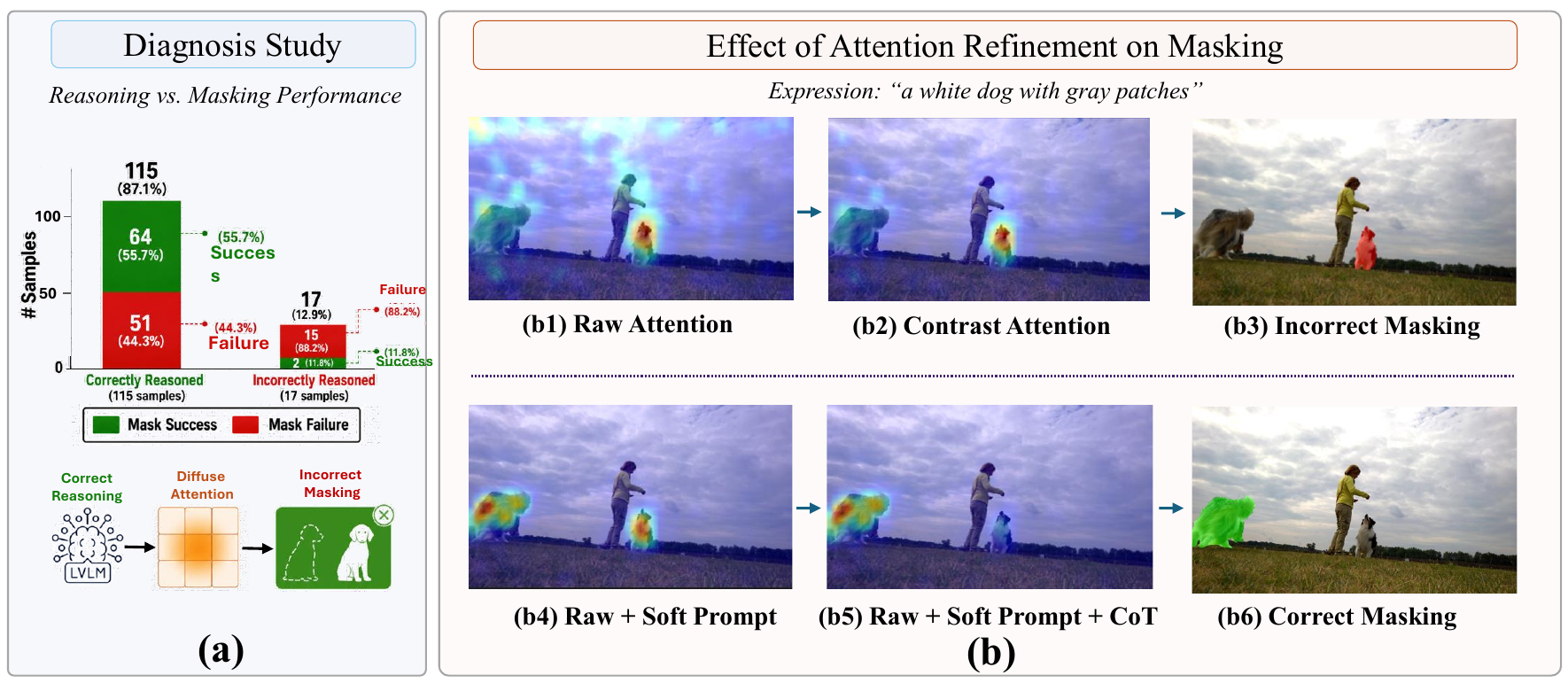}
\caption{
\textbf{(a)} Diagnostic study on reasoning and grounding. While the LVLM often identifies the correct target object, the corresponding attention remains poorly localized, leading to inaccurate masks. \textbf{(b)} Effect of attention refinement on segmentation. Raw and contrast-based attention produce ambiguous localization, while soft prompts and CoT reasoning progressively improve attention quality and segmentation accuracy.
}
\vspace{-0.3cm}
    \label{fig:fig1}
\end{figure}

Recent advances in LVLMs~\cite{liu2023llava, bai2025qwen25vl, chen2024internvl} have shown that their reasoning capabilities implicitly encode localization cues within attention maps. Emerging training-free approaches~\cite{han2026decomposed, kang2025lochead} leverage these attention maps as grounding signals to generate spatial prompts for segmentation models such as SAM2~\cite{ravi2025sam}. However, LVLM attention maps are not optimized for precise localization and are often diffuse, noisy, and affected by attention sinks~\cite{kang2025see}, leading to ambiguous grounding (Fig.~\ref{fig:fig1}(b1)). Although prior methods attempt to mitigate this issue through contrastive prompting~\cite{han2026decomposed}, the resulting attention often remains imprecise and produces inaccurate masks (Fig.~\ref{fig:fig1}(b2)).

To better understand this gap, we conduct a diagnostic study on 132 samples randomly selected from the ReasonVOS~\cite{bai2024one} dataset. For each query, we prompt Qwen2.5-VL \cite{bai2025qwen25vl} to generate distinguishing attributes describing the referred object. We then evaluate the responses through human annotation using a custom-designed GUI interface (see Appendix \ref{sec:A}). The evaluation shows that the model correctly identifies the target object in 115 cases (87.1\%), indicating strong semantic reasoning capability (Fig. \ref{fig:fig1} (a)). However, when applying the state-of-the-art attention-based segmentation method DecAF~\cite{han2026decomposed}, only 55.7\% of these correctly understood cases produce accurate masks, while only 2 out of 17 incorrectly reasoned cases are successfully segmented.
These results reveal a clear discrepancy between reasoning and grounding: correct reasoning is necessary but not sufficient for accurate localization. These results suggest that the main limitation lies in attention localization, motivating the need for explicit attention steering.

To this end, we adopt an input-level attention steering strategy that operates directly at the source of attention formation. Specifically, we prepend a small set of learnable soft prompts to a frozen LVLM, allowing them to participate in self-attention at every layer. By influencing attention throughout the network, these prompts reshape how the generated response token attends to visual tokens, yielding more concentrated and spatially aligned attention maps (Fig.~\ref{fig:fig1}(b4)).
To further resolve ambiguity among visually similar objects, we incorporate a lightweight single-step Chain-of-Thought (CoT) reasoning module. Given the input query, the model first generates discriminative attributes (e.g., color, position, motion), which are appended to the original prompt. This attribute-augmented prompt guides attention toward the correct instance through explicit semantic cues, producing more discriminative attention maps (Fig.~\ref{fig:fig1}(b5)).
Finally, from the refined attention maps, we generate point prompts by selecting high-confidence locations across sampled keyframes. These points are fed to a segmentation model to produce candidate masks, which are propagated into tracklets. We then introduce a correlation-based scoring mechanism that measures consistency between each candidate and the attention signals, enabling robust tracklet selection under challenging conditions.

Importantly, our approach requires no fine-tuning of the LVLM and preserves its pretrained reasoning capabilities, while effectively adapting its attention for accurate and reliable grounding. Building upon these insights, we summarize the primary contributions of our work as follows.
\textbf{First}, we identify attention misalignment as the primary bottleneck in attention-based grounding and show that attention rollout \cite{abnar2020attentionrollout}, traditionally used as a post-hoc visualization tool, can be effectively controlled through input-level conditioning. 
\textbf{Second}, we propose a principled attention steering framework that combines learnable soft prompts for spatial attention concentration with Chain-of-Thought (CoT) reasoning for ambiguity resolution, enabling accurate grounding without modifying LVLM parameters. 
\textbf{Third}, we show that explicit attention steering enables transferable object-centric grounding in LVLMs, yielding consistent cross-dataset generalization without task-specific fine-tuning.

\section{Related work}

\vspace{-0.2cm}

 \paragraph{Large Vision-Language Models.}
LLMs have demonstrated powerful reasoning and cognition capabilities~\cite{brown2020gpt3,dubey2024llama3,yang2024qwen2}, motivating their extension to vision through Large vision-language models (LVLMs). Early LVLMs such as Flamingo~\cite{alayrac2022flamingo}, BLIP-2~\cite{li2023blip2}, and LLaVA~\cite{liu2023llava} established the standard paradigm of connecting a frozen vision encoder~\cite{radford2021clip} to an LLM through a lightweight projection module. This design was later refined by models such as LLaVA-1.5~\cite{liu2024llava15}, InternVL~\cite{chen2024internvl}, Cambrian-1~\cite{tong2024cambrian}, Molmo~\cite{deitke2025molmo}, and InternVL3~\cite{zhu2025internvl3}. These models are built on the Transformer architecture~\cite{vaswani2017attention} and rely on the self-attention mechanism, whose quadratic cost is prohibitive for video inputs. To mitigate this, many video LVLMs compress visual tokens into fixed-length representations via lightweight aggregation modules~\cite{maaz2024videochatgpt,jin2024chatunivi,li2024llamavid,song2024moviechat,shen2025longvu,ma2025llavamini}. Such compression discards fine-grained spatial cues, unlike LLaVA-style designs~\cite{liu2023llava,zhang2025llavavideo} that retain dense patch features through a simple linear projector. More recently, Qwen2-VL~\cite{wang2024qwen2vl}, Qwen2.5-VL~\cite{bai2025qwen25vl}, and Oryx~\cite{liu2025oryx} further advance this direction through improved high-resolution processing while preserving fine-grained spatial details. In this work, we build on these models and explore the localization capability of LVLMs.

\vspace{-0.2cm}
\paragraph{Attention-Guided Grounding in LVLMs.}
A growing body of work observes that self-attention maps inside LVLMs encode rich visual-textual correspondences and can support downstream grounding tasks without retraining. Building on attention rollout~\cite{abnar2020attentionrollout}, VL-SAM~\cite{lin2024vlsam} aggregates attention across heads and layers and uses the resulting maps as prompts for SAM to enable open-ended detection and segmentation, while Loc-Head~\cite{kang2025lochead} shows that only a few frozen attention heads suffice for competitive visual grounding. F-LMM~\cite{wu2025flmm} converts word--pixel correspondences from frozen LVLM attention into segmentation masks, and ControlMLLM~\cite{wu2024controlmllm} guides attention via test-time optimization for training-free visual prompting. ``LVLMs Know Where to Look''~\cite{zhang2025knowwheretolook} further shows that internal attention can reliably localize objects even when the generated textual answer is incorrect.
\vspace{-0.2cm}
\paragraph{Video Reasoning Segmentation.}
Video reasoning segmentation aims to localize and track objects given complex, world-knowledge-dependent text queries~\cite{yan2024visa,bai2024one}, going beyond traditional referring video object segmentation~\cite{khoreva2018video,seo2020urvos,ding2023mevis}. Existing approaches fall into two broad categories. \textit{Training-based methods} couple LVLMs with mask decoders and fine-tune the joint model on large-scale segmentation data: LISA~\cite{lai2024lisa} introduces a \texttt{<SEG>} token bridging an LVLM with SAM~\cite{kirillov2023sam}, GLaMM~\cite{rasheed2024glamm} and PixelLM~\cite{ren2024pixellm} extend it to grounded conversation and multi-target segmentation, while VISA~\cite{yan2024visa}, VideoLISA~\cite{bai2024one}, VRS-HQ~\cite{gong2025vrshq}, and GLUS~\cite{lin2025glus} adapt this paradigm to videos with temporal tokens, keyframe selection, and global--local reasoning. \textit{Training-free methods} instead exploit intrinsic localization signals in frozen LVLMs: VL-SAM~\cite{lin2024vlsam}, Loc-Head~\cite{kang2025lochead}, and F-LMM~\cite{wu2025flmm} convert LVLM attention into prompts or masks in the image domain, while CoT-RVS~\cite{kao2026cotrvs}, SDAM~\cite{zhu2026sdam}, PARSE-VOS~\cite{parse2026vos}, and DecAF~\cite{han2026decomposed} extend this idea to videos via chain-of-thought prompting, spatio-temporal decoupling, hierarchical LLM reasoning, and decomposed attention fusion, respectively. Our approach sits between these two regimes: rather than fully fine-tuning the LVLM or remaining entirely training-free, we perform \emph{lightweight training on a limited amount of data} to steer the LVLM's attention toward referenced objects, preserving its pretrained reasoning ability while delivering accuracy competitive with fully trained systems.

\section{Method}

\label{sec:method}

\begin{figure}
    \centering
    \includegraphics[width=\linewidth]{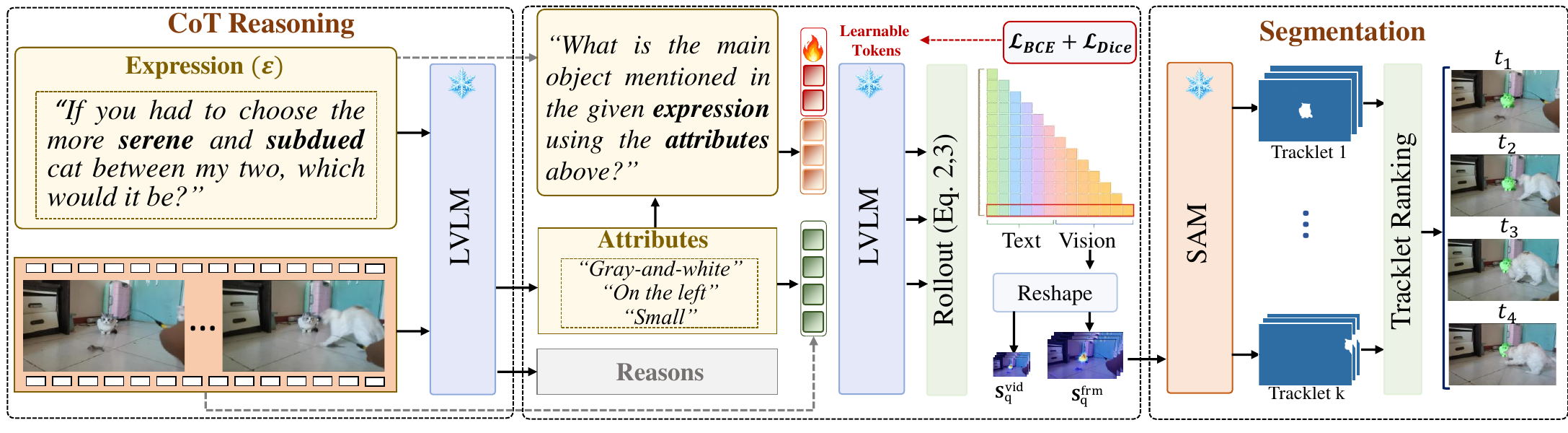}
    \caption{Given a referring expression and video frames, the frozen LVLM first performs CoT reasoning to extract discriminative attributes of the target object. These attributes are incorporated into a prompt for response-token generation, together with learnable soft prompts. Attention rollout from the response token produces spatial attention maps, which are reshaped into frame-level and video-level localization cues. The resulting attention maps are used to guide SAM for mask propagation and candidate tracklet generation, followed by tracklet ranking to obtain the final segmentation mask.}
    \label{fig:fig2}
\end{figure}


\vspace{-0.2cm}

\textbf{Problem Formulation. }We study \emph{video reasoning segmentation}, where the goal is to localize and segment a target object across video frames given a natural language query that may require reasoning.
Formally, we are given a video $\mathcal{V} = \{\mathbf{I}_t\}_{t=1}^{T}$, where each frame $\mathbf{I}_t \in \mathbb{R}^{H \times W \times 3}$, and a natural language expression $\mathcal{E}$ (e.g., ``the person who served the ball''). The objective is to predict a sequence of segmentation masks:
\begin{equation}
\mathcal{M} = \{\mathbf{M}_t\}_{t=1}^{T}, \quad \mathbf{M}_t \in [0,1]^{H \times W},
\end{equation}
where $\mathbf{M}_t$ denotes the spatial extent of the target object at frame $t$.
Unlike standard referring segmentation, the expression $\mathcal{E}$ may require multi-step reasoning, temporal understanding, or relational inference. We assume access to a pretrained large vision-language model (LVLM) and a segmentation foundation model (e.g., SAM2~\cite{ravi2025sam}), both kept frozen. Our goal is to exploit the intrinsic cross-modal attention of the LVLM to derive spatio-temporal grounding signals for accurate segmentation.


\subsection{Response-Token Attention Extraction}

Given a video-expression pair $(\mathcal{V}, \mathcal{E})$, we prompt the frozen LVLM with a concise instruction to identify the main object referred to in $\mathcal{E}$ and respond with a single-word descriptor, which we denote by $q$. We treat this token as the query and analyze its attention over the token sequence to understand how the model grounds the target concept in the visual input.

At transformer layer $\ell$, let $\mathbf{A}^{(\ell)} \in \mathbb{R}^{H \times N \times N}$ denote the attention tensor, where $H$ is the number of heads and $N = N_v + N_t$ is the total number of tokens, consisting of $N_v$ visual tokens and $N_t$ non-visual tokens. Each slice $\mathbf{A}_h^{(\ell)} \in \mathbb{R}^{N \times N}$ represents the pairwise attention weights between tokens for head $h$.
Following \cite{abnar2020attentionrollout}, we first average across heads and incorporate residual connections and compute the attention rollout from layer $L_0$ to $L$ as:

\begin{equation}
\mathbf{R}=\prod_{\ell=L_0}^{L}\tilde{\mathbf{A}}^{(\ell)},
\qquad
\tilde{\mathbf{A}}^{(\ell)}=\tfrac{1}{2}\left(\tfrac{1}{H}\sum_{h=1}^{H}\mathbf{A}_h^{(\ell)}+\mathbf{I}\right).
\end{equation}

where $\mathbf{R} \in \mathbb{R}^{N \times N}$ captures the aggregated token-to-token influence across layers.
Given the response token $q$, let $i_q$ denote its index in the token sequence. The attention corresponding to $q$ is obtained from the rollout matrix $\mathbf{R}$ as:
\begin{equation}
\mathbf{r}_q = \mathbf{R}[i_q, :] \in \mathbb{R}^{1 \times N}, \quad
\mathbf{r}_q^{\text{vis}} = \mathbf{r}_q[:, \mathcal{I}_{\text{vis}}] \in \mathbb{R}^{1 \times N_v},
\end{equation}
where $\mathcal{I}_{\text{vis}}$ denotes the index set of visual tokens. 
{This provides a generic query-conditioned visual attention vector, which can be reshaped according to the visual-token layout to form a spatial grounding map.}

{\textit{\textbf{Dual-Granularity Rollout Attention.} }
The rollout formulation above is agnostic to the visual-token layout. We therefore instantiate it at two complementary granularities to balance spatial precision and temporal context. For the \textit{frame modality}, we uniformly sample $T_f=16$ key frames and compute rollout attention independently for each frame, yielding attention maps $\{\mathbf{S}^{(t)}_q\}_{t=1}^{T_f}$, where $\mathbf{S}^{(t)}_q\in\mathbb{R}^{H_v\times W_v}$. We denote their stack by $\mathbf{S}^{\mathrm{frm}}_q\in\mathbb{R}^{T_f\times H_v\times W_v}$. This modality preserves high spatial resolution and provides precise localization cues. For the \textit{video modality}, we process $T_v=8$ key frames jointly to capture temporal consistency across the sequence. To reduce the computational cost of 3D attention, we apply token merging and spatial downsampling by a factor of $2$, producing a video-level rollout tensor $\mathbf{S}^{\mathrm{vid}}_q\in\mathbb{R}^{T_v\times\tilde{H}_v\times\tilde{W}_v}$, where $\tilde{H}_v=H_v/2$ and $\tilde{W}_v=W_v/2$.}

\subsection{Attention Steering}
The rollout map 
provides grounding cues but is often diffuse, noisy, and prone to attention sinks \cite{kang2025see}, reflecting coarse semantic relevance rather than precise localization ({Fig.~\ref{fig:fig1} (b1)}). To address this, we keep the model fixed and steer attention through input-level conditioning, aiming to \emph{align attention directly to the target object} and obtain more focused and discriminative attention (Fig.~1(b4--b5)). Specifically, we introduce two complementary mechanisms: (1) prompt-based steering to concentrate attention, and (2) reasoning-based steering to resolve ambiguity.

\textit{\textbf{Prompt-Based Steering.}} We introduce a set of learnable prompt tokens to guide the attention formation process while keeping the LVLM frozen. Let $\mathbf{P} \in \mathbb{R}^{N_p \times d}$ denote the $N_p$ learnable prompt embeddings, where $d$ is the embedding dimension. These prompt tokens are appended to the input sequence used for generating the response token. Consequently, the sequence length increases from $N$ to $N + N_p$. This also changes the attention tensors at layer $\ell$ to $\mathbf{A}^{(\ell)} \in \mathbb{R}^{H \times (N+N_p) \times (N+N_p)}$. The resulting attention rollout matrix is therefore given by $\mathbf{R} \in \mathbb{R}^{(N+N_p) \times (N+N_p)}$.

Since the grounding map $\mathbf{S}_q$ is obtained from the rollout matrix $\mathbf{R}$, optimizing the soft prompts $\mathbf{P}$ directly influences $\mathbf{S}_q$. We therefore train $\mathbf{P}$ to make $\mathbf{S}_q$ align with the target object. To this end, we supervise the rollout-based attention ($\mathbf{S}_q$) using ground-truth segmentation masks. We train the soft prompts $\mathbf{P}$ on the training set of Ref-YouTube-VOS dataset ~\cite{seo2020urvos}, which provides pixel-level annotations for referred objects in videos. Let $\hat{\mathbf{G}} \in [0,1]^{H_v \times W_v}$ denote the resized ground-truth mask, and let $\mathbf{S}_q \in [0,1]^{H_v \times W_v}$ denote the normalized rollout attention map for the query token $q$. We optimize the soft prompts using a combination of binary cross-entropy and Dice loss:
\begin{equation}
\mathcal{L} = \mathcal{L}_{\text{BCE}}(\mathbf{S}_q, \hat{\mathbf{G}}) + \mathcal{L}_{\text{Dice}}(\mathbf{S}_q, \hat{\mathbf{G}}).
\end{equation}

The binary cross-entropy term enforces pixel-wise alignment, while the Dice loss encourages overlap between the predicted attention map and the ground-truth mask, improving robustness to class imbalance. Together, these losses guide the soft prompts to produce attention maps that are both localized and structurally consistent with the target object.



\textit{\textbf{Reasoning-Guided  Steering}.} Despite the attention concentration achieved by soft prompts, ambiguity remains a challenge for expressions requiring deeper reasoning. As shown in Fig.~\ref{fig:fig1}(b4), when multiple objects share similar appearance, the attention map may localize an incorrect instance. This typically occurs when the model fails to reason correctly about the expression, indicating that accurate reasoning is necessary for reliable grounding.

To address this, we incorporate a lightweight, single-round CoT reasoning step to guide attention. Given the expression $\mathcal{E}$, we prompt the LVLM (without soft prompts) to briefly reason about the target object and output a small set of distinguishing attributes $a = \{a_1, \ldots, a_K\}$ (e.g., color, position, motion) that help disambiguate it from similar instances; the full prompt template is provided in {Appendix \ref{sec:cot_prompt}}. We append these attributes to the original prompt used for response token generation, forming an attribute-augmented prompt (the full prompt template is provided in Appendix~\ref{sec:query_prompt}). The input sequence is further prepended with the soft prompts $\mathbf{P}$, resulting in a unified prompt that combines $\mathcal{E}$, $a$, and $\mathbf{P}$ (see Fig. \ref{fig:fig2}).

Intuitively, this design improves grounding in two complementary ways. First, the model performs an explicit reasoning step that extracts attributes capturing distinguishing properties of the target object. Second, these attribute tokens are jointly processed with visual tokens and soft prompts, enabling richer interactions across tokens. This enriched context reduces ambiguity by guiding the attention of the query token toward the correct instance, resulting in more accurate and reliable grounding.

\subsection{Mask Tracklets Generation and Selection}



We use the frame-level attention maps
to generate candidate masks, as they provide higher spatial resolution and more precise localization cues compared to the video-level maps. For each key frame $t$, we select the location with the maximum attention value,
\begin{equation}
(x_t, y_t) = \arg\max_{(x,y)} \mathbf{S}_q^{(t)}(x,y),
\end{equation}
as a point prompt and feed it to the segmentation model (SAM2 ~\cite{ravi2025sam}), producing one mask per frame.  This results in $T_f$ candidate masks. To remove redundant or overlapping predictions, we apply non-maximum suppression (NMS) with an IoU threshold of $0.7$.
Finally, we construct tracklets by propagating the selected masks forward and backward across frames using the segmentation model, producing temporally consistent object tracks.

\textbf{\textit{Best Tracklet Selection.}} Given a set of $N_c$ candidate tracklets, we select the most reliable one by measuring their consistency with the attention signals. Intuitively, a correct tracklet should align well with the rollout attention maps, both spatially (frame-level) and temporally (video-level). For each candidate tracklet, we first sample its predicted masks at the selected key frames and stack them into a tensor $\mathbf{M} \in \mathbb{R}^{T_f \times H_v \times W_v}$. We transform these masks into the logit space to better capture correlation in a continuous domain.
We then compute the Pearson correlation between $\mathbf{M}$ and the rollout attention tensors.
{
For the frame modality, we directly compare $\mathbf{M}$ with the frame-level rollout tensor $\mathbf{S}^{\mathrm{frm}}_q\in\mathbb{R}^{T_f\times H_v\times W_v}$. For the video modality, let $\mathbf{S}^{\mathrm{vid}}_q\in\mathbb{R}^{T_v\times\tilde{H}_v\times\tilde{W}_v}$ denote the video-level rollout tensor. We spatially and temporally align the the candidate mask volume $\mathbf{M}$ to the resolution of the video-level attention map using an alignment operator $\mathcal{A}_{\mathrm{vid}}(\cdot)$, yielding
 $\mathbf{M}^{\mathrm{vid}}=\mathcal{A}_{\mathrm{vid}}(\mathbf{M})\in\mathbb{R}^{T_v\times\tilde{H}_v\times\tilde{W}_v}$. The two agreement scores are computed as
\begin{equation}
s_{\mathrm{frm}}=\operatorname{corr}(\mathbf{M},\mathbf{S}^{\mathrm{frm}}_q),
\qquad
s_{\mathrm{vid}}=\operatorname{corr}(\mathbf{M}^{\mathrm{vid}},\mathbf{S}^{\mathrm{vid}}_q),
\end{equation}
where $\operatorname{corr}(\cdot,\cdot)$ denotes the Pearson correlation computed over all elements. We combine the two scores as
$s=\alpha s_{\mathrm{frm}}+(1-\alpha)s_{\mathrm{vid}}$
and select the tracklet with the highest score. 
The selected tracklet is then propagated over all frames and resized to the original image resolution to obtain the predicted mask sequence $\mathcal{M}$.
This design leverages the complementary strengths of the two modalities: frame-level attention provides fine-grained spatial alignment, while video-level attention captures temporal consistency.
}

\section{Experiments}
\label{sec:experiment}


\textbf{Datasets and Evaluation Metrics. } We evaluate on five referring video object segmentation benchmarks, including three conventional referring VOS datasets: Ref-DAVIS~\cite{khoreva2018video}, Ref-YouTube-VOS~\cite{seo2020urvos}, and MeViS~\cite{ding2023mevis}, as well as two reasoning-intensive datasets: ReasonVOS~\cite{bai2024one} and ReVOS~\cite{yan2024visa}. 
ReasonVOS mainly focuses on complex reasoning and compositional understanding, while ReVOS additionally contains challenging reasoning-based referring expressions. 
Following prior works, we report the standard video object segmentation metrics: region similarity ($\mathcal{J}$), contour accuracy ($\mathcal{F}$), and their average ($\mathcal{J}\&\mathcal{F}$).


\textbf{Implementation Details.} We use publicly available LVLM checkpoints, including LLaVA-OV-7B~\cite{li2024llava}, InternVL3-8B~\cite{zhu2025internvl3}, Qwen2VL-7B~\cite{wang2024qwen2vl}, and Qwen2.5VL-7B~\cite{bai2025qwen25vl}. SAM2-hiera-large~\cite{ravi2025sam} is employed as the segmentation model.
Attention rollout starts from the middle transformer layers ($L_0$ for Qwen2.5VL-7B; more details are provided in the {Appendix \ref{sec:d}). We use $64$ learnable soft prompt tokens. The soft prompts are trained only on the training split of Ref-YouTube-VOS ~\cite{seo2020urvos} using supervision from ground-truth masks with BCE and Dice losses, while all LVLM parameters remain frozen during both training and inference. We train the soft prompts using a learning rate of $5\times10^{-4}$ for 6500 steps with an effective batch size of 4 on two NVIDIA A6000 GPUs. The soft prompts are initialized from token embeddings of fixed natural-language seed prompts for the frame and video branches using a repeat-and-truncate strategy, rather than random initialization.
Unless otherwise stated, we use $\alpha=0.3$ for combining frame-level and video-level Pearson correlation scores during tracklet ranking, which is selected based on hyperparameter analysis in the ablation study. 

\subsection{Results}

\begin{table*}[t]
\centering
\caption{
Comparison of LVLM-based text-conditioned VOS methods on Ref-DAVIS ~\cite{khoreva2018video}, ReasonVOS ~\cite{bai2024one}, and ReVOS ~\cite{yan2024visa} benchmarks. The upper gray rows correspond to methods trained end-to-end jointly with the LVLM and segmentation model, while the lower colored rows use frozen LVLMs. Methods marked with $^\star$ are fully training-free.
}
\label{tab:comparison2}

\definecolor{trainrow}{RGB}{230,230,230}
\definecolor{baselineorange}{RGB}{255,238,220}
\definecolor{oursorange}{RGB}{255,214,170}

\definecolor{baselinepurple}{RGB}{238,232,255}
\definecolor{ourspurple}{RGB}{215,200,255}

\definecolor{baselineteal}{RGB}{220,245,245}
\definecolor{oursteal}{RGB}{175,235,235}

\definecolor{baselinegreen}{RGB}{225,245,225}
\definecolor{oursgreen}{RGB}{180,235,180}

\resizebox{\textwidth}{!}{%
\begin{tabular}{llccccccccccccccc}
\toprule

\multirow{2}{*}{Method} & \multirow{2}{*}{LVLM} 
& \multicolumn{3}{c}{Ref-DAVIS}
& \multicolumn{3}{c}{ReasonVOS}
& \multicolumn{3}{c}{ReVOS\textsubscript{(Overall)}}
& \multicolumn{3}{c}{ReVOS\textsubscript{(Referring)}}
& \multicolumn{3}{c}{ReVOS\textsubscript{(Reasoning)}} \\

\cmidrule(lr){3-5}
\cmidrule(lr){6-8}
\cmidrule(lr){9-11}
\cmidrule(lr){12-14}
\cmidrule(lr){15-17}

& 
& $\mathcal{J\&F}$ & $\mathcal{J}$ & $\mathcal{F}$
& $\mathcal{J\&F}$ & $\mathcal{J}$ & $\mathcal{F}$
& $\mathcal{J\&F}$ & $\mathcal{J}$ & $\mathcal{F}$
& $\mathcal{J\&F}$ & $\mathcal{J}$ & $\mathcal{F}$
& $\mathcal{J\&F}$ & $\mathcal{J}$ & $\mathcal{F}$ \\

\midrule

\multicolumn{17}{c}{\cellcolor{trainrow}\textbf{Fully Trained Methods}} \\
\midrule

\rowcolor{trainrow}
LISA\textsubscript{[CVPR'24]}
& LLaVA-7B
& 64.8 & 62.2 & 67.3
& 31.1 & 29.1 & 33.1
& 40.9 & 39.1 & 42.7
& 45.7 & 44.3 & 47.1
& 36.1 & 33.8 & 38.4 \\

\rowcolor{trainrow}
VISA\textsubscript{[ECCV'24]}
& ChatUniVi-7B
& 69.4 & 66.3 & 72.5
& -- & -- & --
& 46.9 & 44.9 & 49.0
& 50.9 & 49.2 & 52.6
& 43.0 & 40.6 & 45.4 \\

\rowcolor{trainrow}
VideoLISA\textsubscript{[NeurIPS'24]}
& LLaVA-Phi-3-V
& 68.8 & 64.9 & 72.7
& 47.5 & 45.1 & 49.9
& -- & -- & --
& -- & -- & --
& -- & -- & -- \\

\rowcolor{trainrow}
GLUS\textsubscript{[CVPR'25]}
& LLaVA-7B
& -- & -- & --
& 49.9 & 47.5 & 52.4
& 54.9 & 52.4 & 57.3
& 58.3 & 56.0 & 60.7
& 51.4 & 48.8 & 53.9 \\

\rowcolor{trainrow}
VRS-HQ\textsubscript{[CVPR'25]}
& ChatUniVi-7B
& 76.0 & 72.6 & 79.4
& -- & -- & --
& 59.1 & 56.6 & 61.6
& 62.1 & 59.8 & 64.5
& 56.1 & 53.5 & 58.7 \\

\rowcolor{trainrow}
Veason-R1\textsubscript{[arxiv'25.08]}
& Qwen2.5VL-7B
& -- & -- & --
& 59.9 & 56.0 & 63.8
& 61.3 & 58.2 & 64.4
& 63.6 & 60.7 & 66.5
& 59.0 & 55.8 & 62.2 \\

\midrule

\multicolumn{17}{c}{\cellcolor{bluerow}\textbf{Frozen-LVLM Methods}} \\
\midrule


\rowcolor{baselineorange}
Loc-Head*\textsubscript{[CVPR'25]}
& LLaVA-7B
& 56.3 & 52.1 & 60.5
& 33.6 & 29.3 & 38.0
& 32.5 & 28.2 & 36.9
& 36.9 & 32.5 & 41.3
& 28.1 & 23.8 & 32.5 \\

\rowcolor{baselineorange}
DecAF*\textsubscript{[ICLR'26]}
& LLaVA-OV-7B
& 59.4 & 54.8 & 64.0
& 52.8 & 49.3 & 56.3
& 40.0 & 35.8 & 44.1
& 43.4 & 39.1 & 47.6
& 36.6 & 32.6 & 40.7 \\

\rowcolor{oursorange}
SteerSeg\textsubscript{[Ours]}
& LLaVA-OV-7B
& \textbf{70.0} & \textbf{65.7} & \textbf{74.3}
& \textbf{58.6} & \textbf{55.7} & \textbf{61.5}
& \textbf{49.2} & \textbf{45.6} & \textbf{52.8}
& \textbf{51.9} & \textbf{48.4} & \textbf{55.5}
& \textbf{47.0} & \textbf{43.4} & \textbf{50.6} \\

\midrule


\rowcolor{baselinepurple}
Loc-Head*\textsubscript{[CVPR'25]}
& InternVL3-8B
& 66.3 & 62.4 & 70.2
& 44.3 & 41.0 & 47.5 
& 43.7 & 39.9 & 47.5
& 46.7 & 42.9 & 50.6
& 43.2 & 39.5 & 46.8 \\

\rowcolor{baselinepurple}
DecAF*\textsubscript{[ICLR'26]}
& InternVL3-8B
& 62.8 & 56.9 & 68.6
& 58.9 & 55.1 & 62.7
& 47.4 & 43.7 & 51.2
& 51.7 & 47.9 & 55.5
& 43.2 & 39.5 & 46.8 \\

\rowcolor{ourspurple}
SteerSeg\textsubscript{[Ours]}
& InternVL3-8B
& \textbf{66.1} & \textbf{62.0} & \textbf{70.1}
& \textbf{63.3} & \textbf{60.6} & \textbf{66.1}
& \textbf{52.5} & \textbf{48.8} & \textbf{56.2}
& \textbf{55.6} & \textbf{51.9} & \textbf{59.3}
& \textbf{49.5} & \textbf{45.8} & \textbf{53.1} \\

\midrule


\rowcolor{baselineteal}
Loc-Head*\textsubscript{[CVPR'25]}
& Qwen2VL-7B
& 61.9 & 58.0 & 65.8
& 34.0 & 31.8 & 36.2
& 44.0 & 40.8 & 47.2
& 52.7 & 49.1 & 56.2
& 35.4 & 32.6 & 38.2 \\

\rowcolor{baselineteal}
DecAF*\textsubscript{[ICLR'26]}
& Qwen2VL-7B
& 64.1 & 59.4 & 68.9
& 52.5 & 49.0 & 56.0
& 45.3 & 41.6 & 49.0
& 52.7 & 48.9 & 56.4
& 37.9 & 34.3 & 41.5 \\

\rowcolor{oursteal}
SteerSeg\textsubscript{[Ours]}
& Qwen2VL-7B
& \textbf{77.8} & \textbf{74.2} & \textbf{81.4}
& \textbf{63.6} & \textbf{60.8} & \textbf{66.4}
& \textbf{53.8} & \textbf{50.4} & \textbf{57.1}
& \textbf{59.2} & \textbf{56.2} & \textbf{62.3}
& \textbf{48.3} & \textbf{44.6} & \textbf{52.0} \\

\midrule


\rowcolor{baselinegreen}
Loc-Head*\textsubscript{[CVPR'25]}
& Qwen2.5VL-7B
& 64.6 & 60.2 & 68.9
& 41.1 & 37.9 & 44.3
& 47.0 & 43.3 & 50.7
& 53.1 & 49.3 & 56.9
& 40.8 & 37.2 & 44.4 \\

\rowcolor{baselinegreen}
DecAF*\textsubscript{[ICLR'26]}
& Qwen2.5VL-7B
& 75.2 & 70.9 & 79.5
& 63.9 & 60.5 & 67.2
& 54.2 & 50.1 & 58.2
& 58.7 & 54.8 & 62.6
& 49.7 & 45.4 & 53.9 \\

\rowcolor{oursgreen}
SteerSeg\textsubscript{[Ours]}
& Qwen2.5VL-7B
& \textbf{81.4} & \textbf{78.0} & \textbf{84.8}
& \textbf{65.9} & \textbf{63.1} & \textbf{68.7}
& \textbf{56.6} & \textbf{53.5} & \textbf{59.8}
& \textbf{61.1} & \textbf{58.2} & \textbf{63.9}
& \textbf{52.4} & \textbf{48.9} & \textbf{55.9} \\

\bottomrule
\end{tabular}
}
\end{table*}

\paragraph{Comparison with Existing Methods.}

Tab.~\ref{tab:comparison2} compares SteerSeg with both fully trained and frozen-LVLM text-conditioned VOS approaches across multiple LVLM backbones and benchmarks. The upper gray rows (LISA \cite{lai2024lisa}, VISA \cite{yan2024visa}, VideoLISA \cite{bai2024one}, GLUS \cite{lin2025glus}, VRS-HQ \cite{gong2025vrshq}, Veason-R1 \cite{gong2025reinforcing}) correspond to methods that jointly optimize the LVLM and segmentation model in an end-to-end cascade training pipeline, typically requiring extensive training on large-scale datasets. In contrast, the lower rows use frozen LVLMs, where methods marked with $^\star$ are fully training-free. Our method occupies a lightweight middle ground: both the LVLM and SAM2 remain completely frozen, and only a small set of soft prompt tokens is learned.

Compared with frozen-LVLM methods such as Loc-Head$^\star$ \cite{kang2025lochead} and DecAF$^\star$ \cite{han2026decomposed}, SteerSeg consistently achieves strong improvements across datasets and backbones. The gains are especially notable on reasoning-intensive benchmarks such as ReasonVOS and ReVOS (Reasoning), demonstrating the effectiveness of attention steering and CoT-guided attribute conditioning for resolving ambiguity among visually similar objects. For instance, with Qwen2.5VL-7B, SteerSeg improves over DecAF$^\star$ by $6.2$ points on Ref-DAVIS, $2.0$ points on ReasonVOS, and $2.7$ points on ReVOS (Reasoning) in terms of $\mathcal{J\&F}$.

Despite freezing both the LVLM and segmentation model, SteerSeg also achieves performance competitive with or superior to fully trained approaches. On Ref-DAVIS, SteerSeg with Qwen2.5VL-7B achieves $81.4$ $\mathcal{J\&F}$, surpassing VRS-HQ by a large margin. Similar trends are observed on ReasonVOS and ReVOS, suggesting that improving attention alignment can be more effective than expensive end-to-end cascade training.

Another key observation is the strong consistency across different LVLM backbones. SteerSeg consistently improves localization quality for LLaVA-OV, InternVL3, Qwen2VL, and Qwen2.5VL. Moreover, although the soft prompts are trained only on Ref-YouTube-VOS, the method generalizes well to unseen datasets and reasoning-intensive benchmarks, indicating that the proposed attention steering mechanism learns transferable alignment behavior rather than dataset-specific heuristics.

\paragraph{Generalization to Additional Datasets.}

Tab.~\ref{tab:comparison3} further evaluates SteerSeg on MeViS ~\cite{ding2023mevis} and Ref-YouTube-VOS  ~\cite{seo2020urvos} to study its generalization across datasets and LVLM backbones. Similar to previous experiments, the upper rows correspond to fully trained approaches, while the lower rows correspond to frozen-LVLM methods.

Compared with existing frozen-LVLM methods such as Loc-Head$^\star$ and DecAF$^\star$, SteerSeg consistently achieves substantial improvements on both datasets. The best SteerSeg variant achieves $53.1$ $\mathcal{J\&F}$ on MeViS and $67.9$ on Ref-YouTube-VOS, exceeding DecAF$^\star$ by $5.0$ and $8.0$ points, respectively. These results further validate the effectiveness of the proposed attention steering mechanism for improving localization quality.

Despite freezing both the LVLM and SAM2, SteerSeg also achieves performance competitive with or superior to several fully trained approaches. On MeViS, SteerSeg with Qwen2VL-7B achieves the best overall performance, while on Ref-YouTube-VOS it remains competitive with recent end-to-end trained methods without requiring expensive cascade optimization.

\begin{table*}[!t]
\centering

\begin{minipage}[t]{0.65\textwidth}
\centering
\tiny
\caption{Comparison on additional datasets.}
\label{tab:comparison3}

\resizebox{\linewidth}{!}{%
\begin{tabular}{llcccccc}
\toprule
\multirow{2}{*}{Method} & \multirow{2}{*}{LVLM} & \multicolumn{3}{c}{MeViS} & \multicolumn{3}{c}{Ref-YouTube-VOS} \\
\cmidrule(lr){3-5} \cmidrule(lr){6-8}
& & $\mathcal{J\&F}$ & $\mathcal{J}$ & $\mathcal{F}$ & $\mathcal{J\&F}$ & $\mathcal{J}$ & $\mathcal{F}$ \\
\midrule

\rowcolor{grayrow}
VISA\textsubscript{[ECCV'24]}         & Chat-UniVi-7B  & 44.5 & 41.8 & 47.1 & 61.5 & \textbf{59.8} & 63.2 \\
\rowcolor{grayrow}
VideoLISA\textsubscript{[NeurIPS'24]} & LLaVA-Phi-3-V  & 44.4 & 41.3 & 47.6 & 63.7 & 61.7 & 65.7 \\
\rowcolor{grayrow}
GLUS\textsubscript{[CVPR'25]}         & LLaVA-7B       & 51.3 & \textbf{48.5} & 54.2 & 67.3 & 65.5 & 69.0 \\
\rowcolor{grayrow}
VRS-HQ\textsubscript{[CVPR'25]}       & Chat-UniVi-7B  & 50.6 & 47.6 & 53.7 & \textbf{70.4} & \textbf{68.3} & \textbf{72.5} \\
\rowcolor{grayrow}
Veason-R1\textsubscript{[arxiv'25.08]}& Qwen2.5VL-7B   & \textbf{52.2} & 48.4 & \textbf{56.0} & -- & -- & -- \\

\midrule

\rowcolor{orangrow}
Loc-Head*\textsubscript{[CVPR'25]} & Qwen2.5VL-7B & 39.4 & 35.2 & 43.6 & 51.0 & 46.8 & 55.2 \\
\rowcolor{orangrow}
DecAF*\textsubscript{[ICLR'26]} & Qwen2.5VL-7B & 48.1 & 44.0 & 52.1 & 59.9 & 56.2 & 63.5 \\

\midrule

\rowcolor{greenrow}
SteerSeg\textsubscript{[Ours]} & LLaVA-OV-7B & 47.8 & 43.8 & 51.9 & 58.0 & 55.4 & 60.6 \\
\rowcolor{greenrow}
SteerSeg\textsubscript{[Ours]} & InternVL3-8B & 48.2 & 43.8 & 52.5 & 56.4 & 53.4 & 59.4 \\
\rowcolor{greenrow}
SteerSeg\textsubscript{[Ours]} & Qwen2VL-7B & \textbf{53.1} & \textbf{49.4} & \textbf{56.8} & \textbf{67.9} & \textbf{65.9} & \textbf{70.0} \\
\rowcolor{greenrow}
SteerSeg\textsubscript{[Ours]} & Qwen2.5VL-7B & 51.5 & 47.8 & 55.2 & 67.2 & 65.3 & 69.1 \\

\bottomrule
\end{tabular}
}

\end{minipage}
\hspace{0.01\textwidth}
\vrule
\hspace{0.01\textwidth}
\begin{minipage}[t]{0.30\textwidth}
\centering

\vspace{0.1cm}

\includegraphics[width=\linewidth]{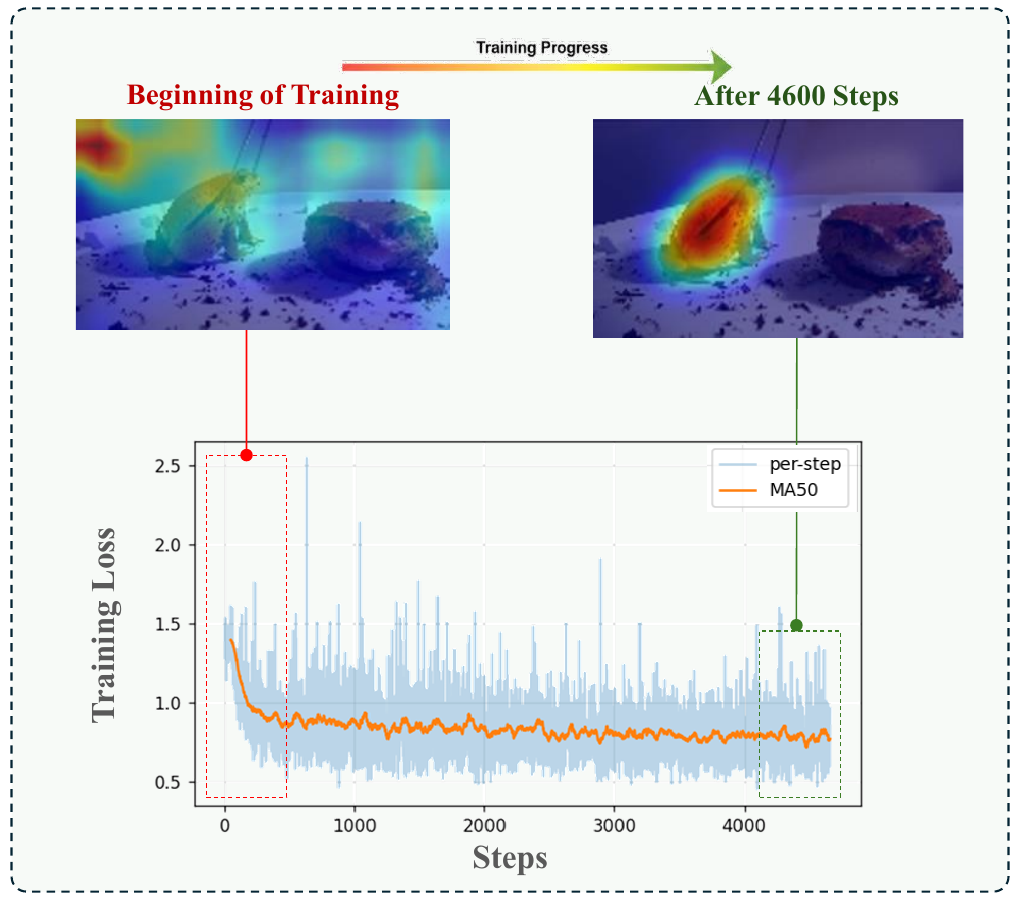}

\captionof{figure}{Attention alignment and training loss evolution.}
\label{fig:training-loss}

\end{minipage}

\end{table*}

\begin{figure}
    \centering
    \includegraphics[width=\linewidth]{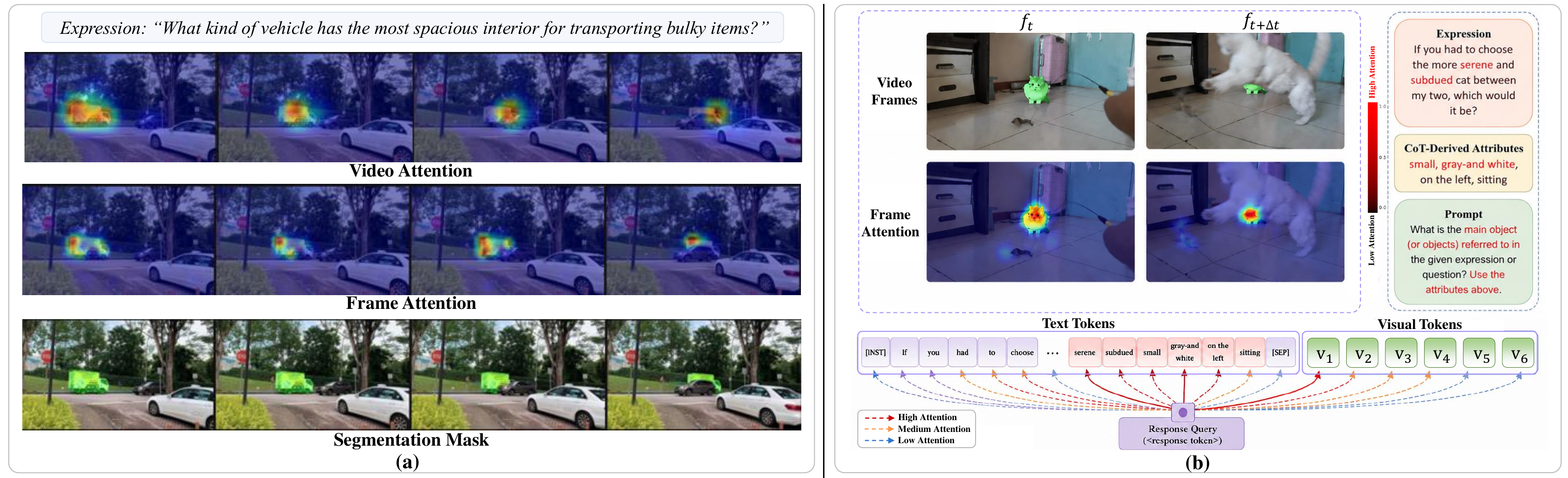}
   \caption{Qualitative visualization of the proposed framework. (a) Video-level attention, frame-level attention, and the resulting segmentation masks across sampled frames. (b) Response-token attention visualization showing interactions between visual tokens and CoT-derived attributes, where red indicates higher attention weights.}
    \label{fig:fig4}
    \vspace{-0.3cm}
\end{figure}


\subsection{Qualitative Analysis}

Fig.~\ref{fig:training-loss} qualitatively illustrates the training dynamics of the proposed attention steering mechanism through both the training loss evolution and the resulting attention maps. As training progresses, the loss steadily decreases while the attention becomes significantly more focused and aligned with the target object. Notably, this example is drawn from a cross-dataset setting, demonstrating that the learned soft prompts generalize beyond the training distribution and still produce coherent object-centric grounding. Fig.~\ref{fig:fig4}(a) presents qualitative examples of the proposed framework. The video-level branch provides temporally consistent localization, while the frame-level branch preserves finer spatial details. Their combination produces accurate and stable segmentation masks under challenging appearance and motion variations. Furthermore, Fig.~\ref{fig:fig4}(b) visualizes the response-token attention mechanism, where the response query attends jointly to visual tokens and CoT-derived attributes. Higher attention weights are highlighted in red, showing that the response token increasingly focuses on semantically relevant attributes and object-centric visual cues.

\vspace{-0.2cm}

\section{Ablation Study}
\label{sec:ablation}



\textbf{Effect of Individual Components.} Table~\ref{tab:ablation_components} analyzes the contribution of CoT-derived attributes and learnable soft prompts. Using CoT alone provides only a modest improvement over the raw attention baseline, suggesting that reasoning alone is not the primary bottleneck in grounding quality. This observation is consistent with our earlier diagnostic analysis, which indicated that the main limitation arises from weak attention alignment rather than missing semantic reasoning. In contrast, introducing learnable soft prompts leads to a significant improvement in both segmentation performance and attention alignment. Specifically, we report the average Pearson correlation between the rollout attention tensors and the resized ground-truth masks across key frames for both the frame-level and video-level modalities over the entire dataset. Combining both components yields the best performance across all datasets and metrics, with the correlation trends remaining consistent with the segmentation results. Overall, the results suggest that CoT guidance is beneficial but not sufficient by itself, while attention steering through soft prompts plays the dominant role.

\begin{table*}[t]
\centering

\begin{minipage}[t]{0.58\textwidth}
\centering
\small
\setlength{\tabcolsep}{5pt}

\begin{tabular}{cccc@{\hspace{10pt}}cc}
\toprule

\multicolumn{2}{c}{\textbf{Components}}
& \multicolumn{2}{c}{\textbf{J\&F}}
& \multicolumn{2}{c}{\textbf{Correlation}} \\

\cmidrule(r){1-2}
\cmidrule(r){3-4}
\cmidrule(l){5-6}

\textbf{Soft}
& \textbf{CoT}
& \textbf{DAVIS}
& \textbf{ReasonVOS}
& \textbf{DAVIS}
& \textbf{ReasonVOS} \\

\midrule

\ding{55} & \ding{55} & 65.7 & 60.6 & 0.19 & 0.25 \\
\ding{55} & \checkmark & 66.6 & 63.6 & 0.21 & 0.30 \\
\checkmark & \ding{55} & 80.9 & 64.1 & 0.41 & 0.44 \\
\checkmark & \checkmark & \textbf{81.4} & \textbf{65.9} & \textbf{0.47} & \textbf{0.45} \\

\bottomrule
\end{tabular}

\caption{Effect of different components on segmentation performance and attention alignment.}

\label{tab:ablation_components}

\end{minipage}
\hfill
\begin{minipage}[t]{0.38\textwidth}
\centering
\small
\renewcommand{\arraystretch}{0.92}

\begin{tabular}{ccc}
\toprule
\textbf{$N_p$} & \textbf{DAVIS} & \textbf{ReasonVOS} \\
\midrule
4   & 78.3 & 63.8 \\
8   & 80.4 & 65.1 \\
16  & 80.6 & 65.4 \\
32  & 80.8 & 65.5 \\
\textbf{64}  & \textbf{81.4} & \textbf{65.9} \\
128 & 81.3 & 65.9 \\
\bottomrule
\end{tabular}

\caption{Effect of the number of soft prompt tokens $N_p$.}
\label{tab:soft_token_ablation}

\vspace{0.4cm}

\end{minipage}

\end{table*}

\textbf{Effect of the Number of Soft Prompt Tokens.} We analyze the effect of the number of learnable soft prompt tokens $N_p$ on segmentation performance. As shown in Table~\ref{tab:soft_token_ablation}, increasing the number of soft tokens leads to consistent but relatively modest improvements on both DAVIS and ReasonVOS. Performance saturates quickly after small prompt sizes, indicating that the proposed attention steering mechanism is not highly sensitive to the exact number of soft tokens. The best performance is achieved at $N_p=64$, which is used in all experiments.

\textbf{Effect of fusion weight $\alpha$ on different query types.} We further analyze the effect of the fusion weight $\alpha$ on different query types in ReVOS. Comparing the two extremes, frame-only attention ($\alpha=1$) performs better on referring queries, highlighting the importance of spatial precision, whereas video-only attention ($\alpha=0$) performs better on reasoning queries, emphasizing the role of temporal context. Nevertheless, the best performance is consistently achieved at an intermediate value ($\alpha \approx 0.3$), showing that neither modality alone is sufficient. Combining frame-level precision with video-level temporal consistency yields the most effective grounding across all query types.

\begin{figure}[!t]
    \centering
    \includegraphics[width=\linewidth]{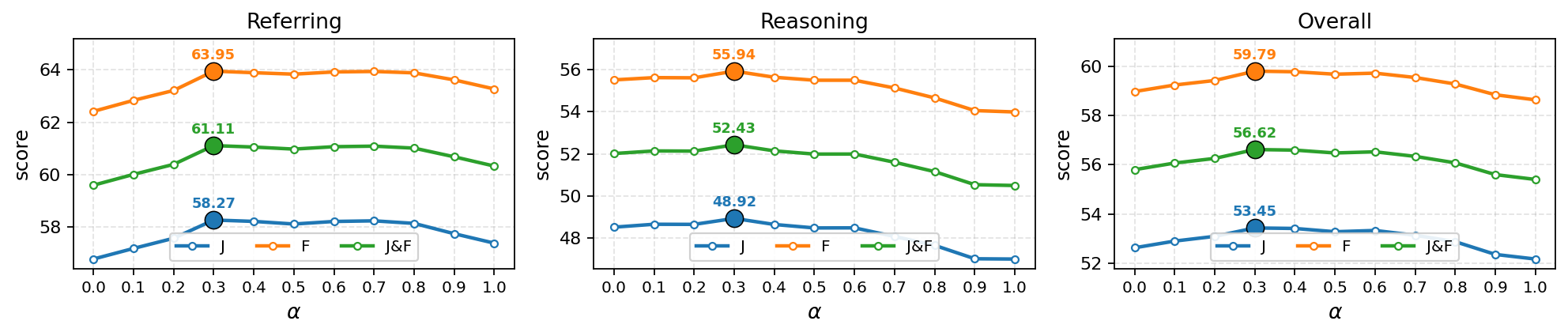}
    \caption{
    Effect of fusion weight $\alpha$ on ReVOS. Frame-only attention ($\alpha=1$) is better for referring queries, while video-only ($\alpha=0$) benefits reasoning tasks requiring temporal context. Performance peaks at $\alpha \approx 0.3$, showing the advantage of combining both.
    }
    \label{fig:placeholder}
\end{figure}





\vspace{-0.1cm}

\section{Conclusion}

We presented SteerSeg, a lightweight attention steering framework for text-conditioned video object segmentation using frozen LVLMs. We showed that the main limitation of attention-based grounding is not the lack of semantic understanding, but the misalignment and diffusion of attention maps. To address this, SteerSeg combines learnable soft prompts for attention concentration with a lightweight CoT-guided attribute extraction mechanism for ambiguity resolution. The resulting attention maps provide more accurate and discriminative localization cues, which are further integrated with SAM2 for robust tracklet generation and segmentation.
Extensive experiments across multiple benchmarks and LVLM backbones demonstrate that SteerSeg consistently improves grounding quality and achieves performance competitive with or superior to fully trained approaches, despite freezing both the LVLM and segmentation model. Moreover, the strong cross-dataset generalization suggests that the proposed attention steering mechanism learns transferable attention alignment behavior rather than dataset-specific heuristics.

\vspace{-0.1cm}

\section{Limitations}
A limitation of our method is that objects appearing only briefly may be missed if they are not included in the sampled key frames, which could be alleviated by adaptive temporal sampling strategies. A further limitation is that the fusion weight is selected empirically and may vary across datasets and LVLM backbones.

\bibliographystyle{abbrvnat}
\bibliography{refs}

\newpage

\appendix

{\Large\textbf{Appendix}}

\section{Diagnostic Study and Annotation Interface}
\label{sec:A}
To better understand the relationship between semantic reasoning and spatial grounding in LVLMs, we conducted a diagnostic study on the ReasonVOS ~\cite{bai2024one} dataset. The goal of this study was to evaluate whether the LVLM can correctly identify the referred object at a semantic level, independent of the quality of the resulting attention maps or segmentation masks.

For each sampled video-expression pair, we prompted Qwen2.5VL-7B \cite{bai2025qwen25vl} to generate a set of distinguishing attributes describing the referred object, including appearance, position, and motion-related cues. We then manually evaluated whether the generated attributes correctly corresponded to the target object referred to in the expression.

To facilitate efficient and consistent annotation, we developed a lightweight graphical user interface, shown in Fig.~\ref{fig:appendix_gui}. The interface displays the sampled video clip as an animated GIF together with the referring expression and the attributes generated by the LVLM. Annotators were asked to mark each sample as \texttt{True} if the generated attributes correctly identified the referred object, or \texttt{False} otherwise. A \texttt{Skip} option was also provided for ambiguous cases.

The GUI additionally tracks annotation progress and allows rapid navigation between samples. Using this interface, we annotated 132 randomly sampled examples from ReasonVOS, which formed the basis of the diagnostic analysis presented in the main paper.

\begin{figure}[!h]
    \centering
    \includegraphics[width=\linewidth]{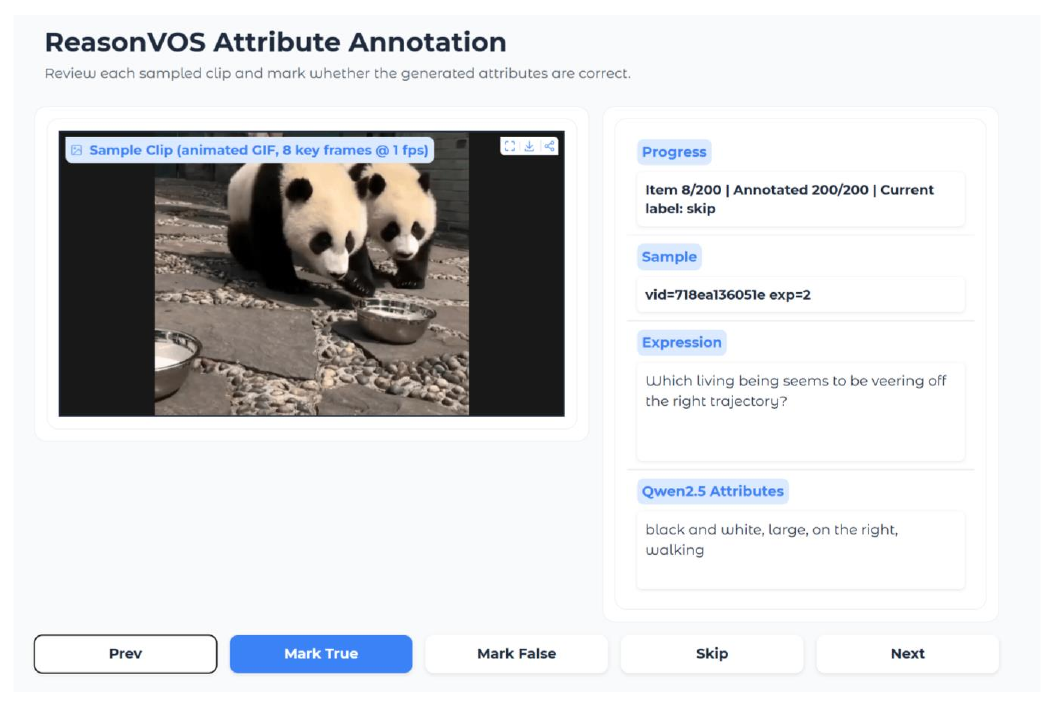}
    \caption{GUI interface used for human evaluation of the generated reasoning attributes on ReasonVOS samples.}
    \label{fig:appendix_gui}
\end{figure}

\section{CoT Prompt for Attribute Extraction}
\label{sec:cot_prompt}

\begin{tcolorbox}[colback=gray!5, colframe=black!80, title=CoT Prompt for Attribute Extraction]
\small
\textbf{Expression:} \{sent\} 

\vspace{0.2cm}
First, briefly reason about which object in the scene best satisfies this expression — consider the role, function, or intent the expression implies, not just visual similarity to the words. Then list distinguishing attributes (color, size, position, shape, motion) of THAT object only.
\vspace{0.2cm}

\textbf{Respond in EXACTLY this format, nothing else:}

\textbf{Reasoning:} \textit{<one or two short sentences>}

\textbf{Attributes:} \textit{<comma-separated list, max $\sim$10 words, e.g., `large, white, on the right, parked'>}
\end{tcolorbox}

During inference, we employ a lightweight Chain-of-Thought (CoT) prompting step to extract discriminative attributes that guide attention. Given an input expression $\mathcal{E}$, the LVLM is prompted to first reason about the target object and then produce a concise set of attributes describing it.

The prompt is designed to encourage the model to move beyond superficial keyword matching and instead identify the target object based on its semantic role and context within the scene. This is particularly important for reasoning-based queries, where the correct object may not be directly described by appearance alone.

The extracted attributes serve as a compact and structured representation of the target, which is appended to the original expression and used together with learnable soft prompts to steer attention. Enforcing a strict output format improves the stability of the extracted attributes and reduces noisy or irrelevant responses, leading to more reliable attention guidance.

\section{Query Prompt for Attention Extraction}
\label{sec:query_prompt}

\begin{tcolorbox}[colback=gray!5, colframe=black!80, title=Query Token Extraction Prompt]
\small
\textbf{Expression:} \{sent\} 

\vspace{0.2cm}

Distinguishing attributes of the target: \{attrs\}.

\vspace{0.2cm}

What is the main object (or objects) referred to in the given expression or question?

Use the attributes above to disambiguate from other similar objects. Respond with a single word (e.g., `cat', `person', `dog') that best describes the target object(s).
\end{tcolorbox}

After obtaining discriminative attributes via the CoT stage (Sec.~\ref{sec:cot_prompt}), we prompt the LVLM to generate a single query token that represents the target object. This token is subsequently used to extract attention maps via attention rollout.

This prompt is designed to produce a concise semantic representation of the target object in the form of a single word. By incorporating the CoT-derived attributes, the model is encouraged to resolve ambiguities and select the correct object category even in the presence of visually similar instances.

The generated token serves as the query in the cross-modal attention mechanism. Specifically, we extract the attention between this token and all visual tokens, and propagate it through attention rollout to obtain a spatio-temporal attention map. This design ensures that the attention map is explicitly grounded in the model’s semantic understanding of the target.

Constraining the output to a single word improves stability and consistency across samples, and prevents the model from generating verbose or ambiguous descriptions that would degrade the quality of the extracted attention.

\section{Implementation Details}
\label{sec:d}

Table~\ref{tab:supp_hparams} summarizes the main hyperparameters used for soft-prompt training, attention-rollout extraction, and SAM2-based mask propagation. Unless otherwise stated, all settings are shared across different LVLM backbones.

\begin{table}[t]
\centering
\small
\setlength{\tabcolsep}{10pt}
\renewcommand{\arraystretch}{1.1}

\begin{tabular}{p{0.45\linewidth} p{0.38\linewidth}}
\toprule
\textbf{Hyperparameter} & \textbf{Value} \\
\midrule

Soft prompt tokens $N_p$ & 64 (canonical) \\
Soft prompt learning rate & $5\times10^{-4}$ \\
Training steps & 6500 \\
Effective batch size & 4 \\
Optimizer & AdamW \\
Learning-rate schedule & Cosine + warmup \\
Training loss & BCE + Dice \\

\midrule

Input frames & 16 \\
Rollout layers & 14--27 \\
Video rollout frames $T_v$ & 8 \\
Frame selection strategy & Uniform sampling \\

\midrule

Point selection strategy & Argmax \\
SAM2 backbone & \texttt{sam2\_hiera\_large} \\
Fusion metric & Pearson correlation \\
Best fusion weight $\alpha$ & 0.3 (default) \\

\bottomrule
\end{tabular}
\vspace{0.1cm}
\caption{Hyperparameters used for soft-prompt training, rollout extraction, and SAM2 propagation.}
\label{tab:supp_hparams}
\end{table}

\section{Qualitative results}
\label{sec:query_qualitive}

\begin{figure}[htbp]
    \centering
    \includegraphics[width=\linewidth]{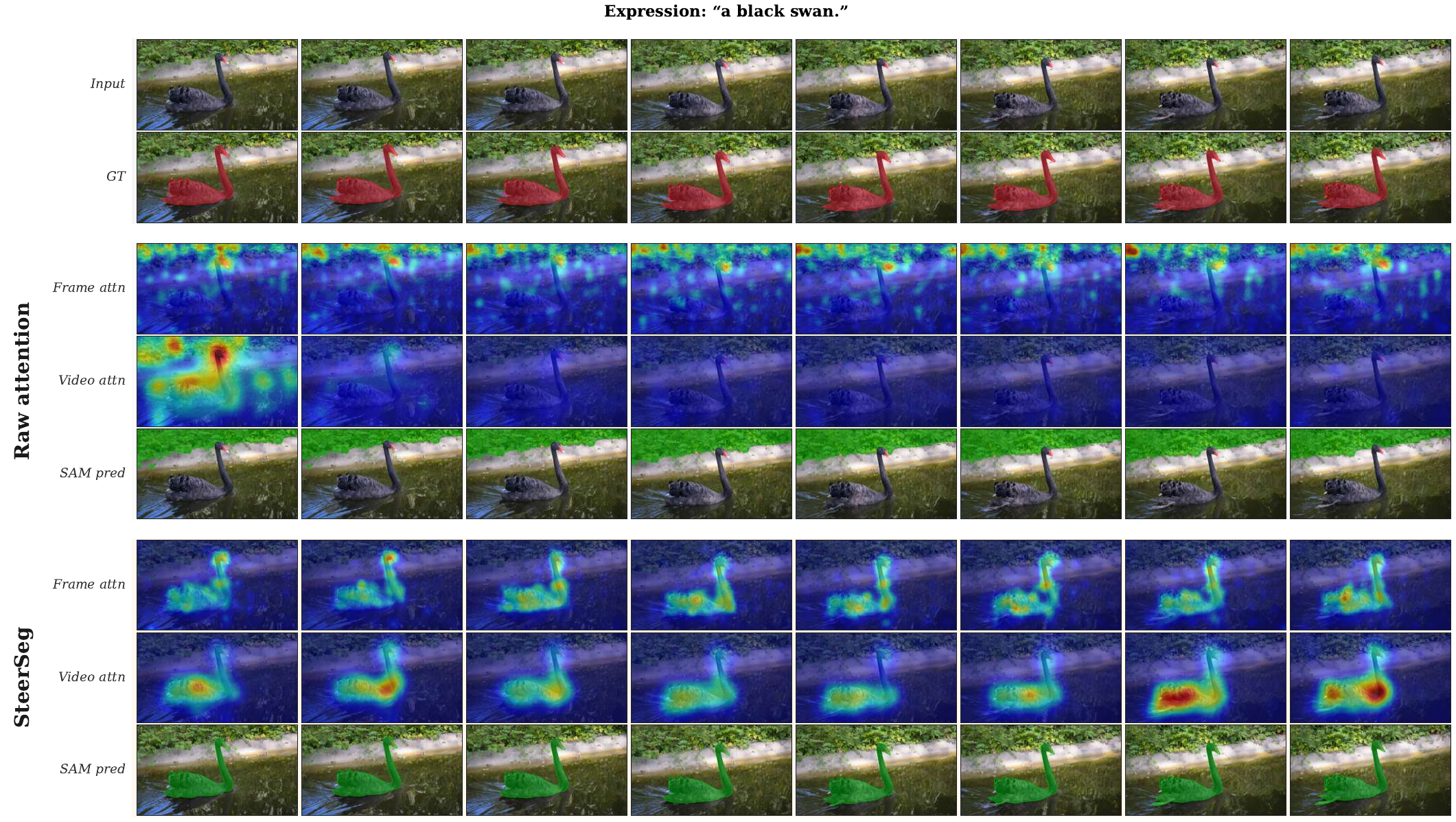}
\end{figure}

\begin{figure}[htbp]
    \centering
    \includegraphics[width=\linewidth]{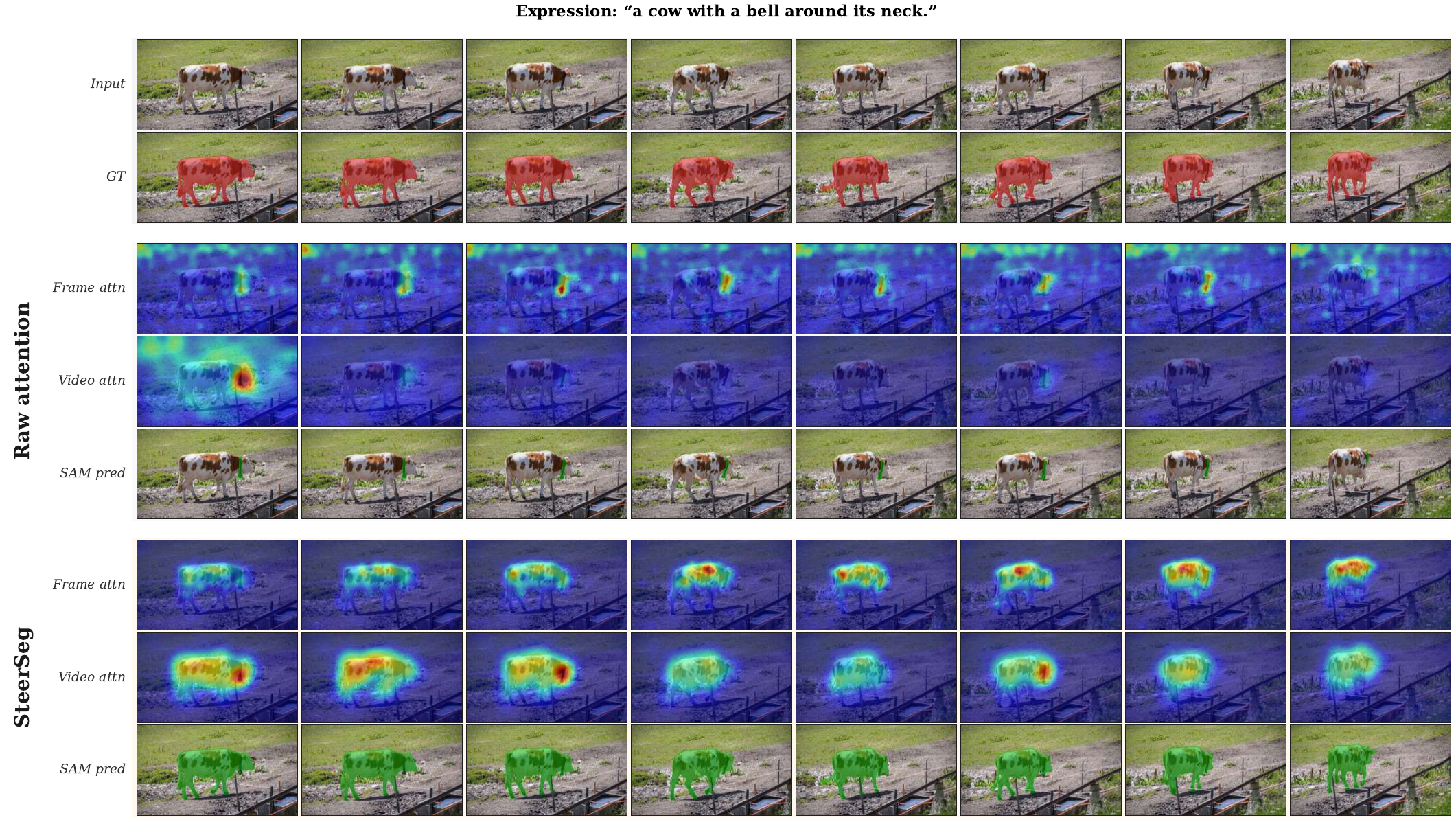}
\end{figure}

\begin{figure}[htbp]
    \centering
    \includegraphics[width=\linewidth]{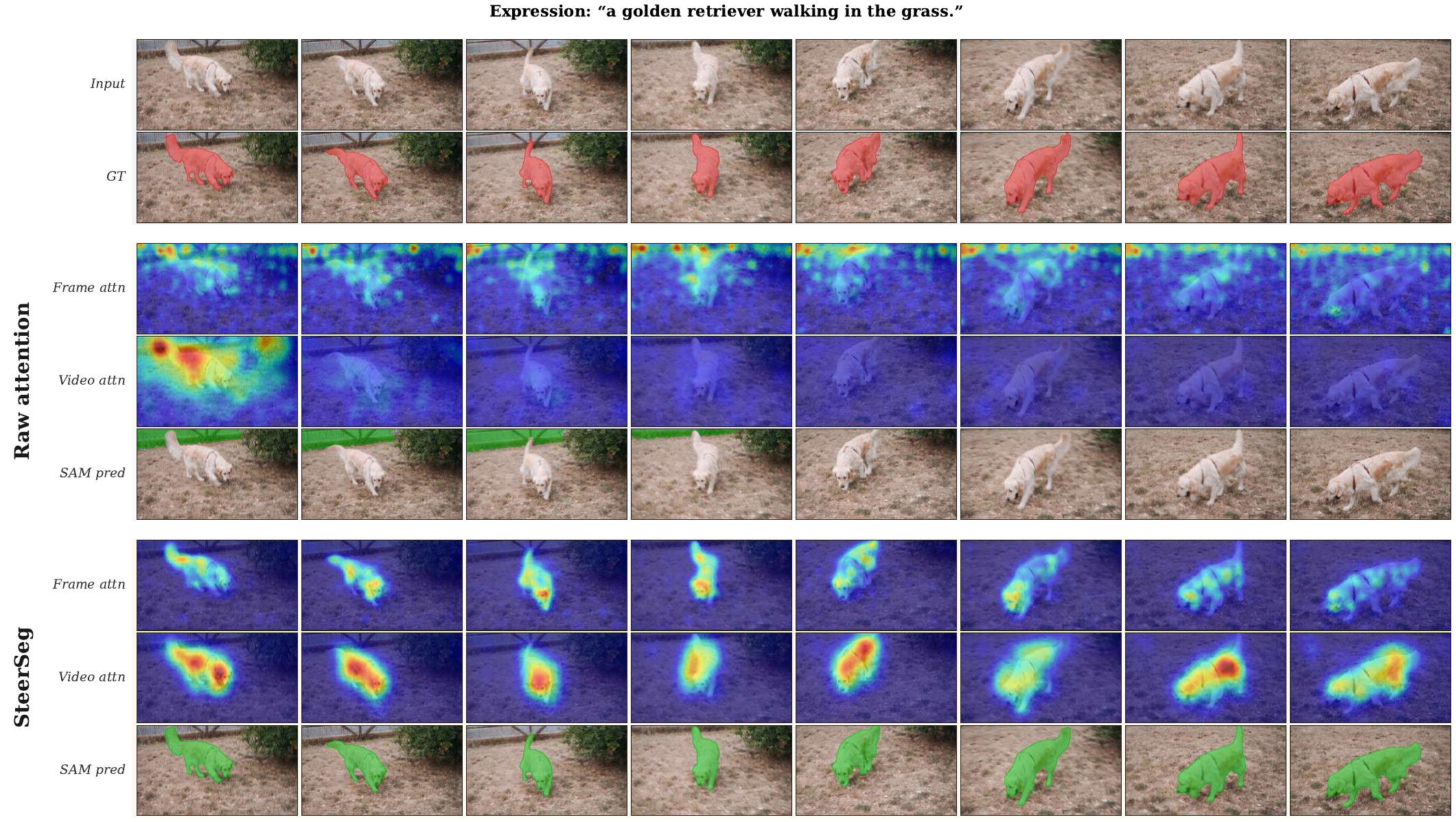}
\end{figure}

\begin{figure}[htbp]
    \centering
    \includegraphics[width=\linewidth]{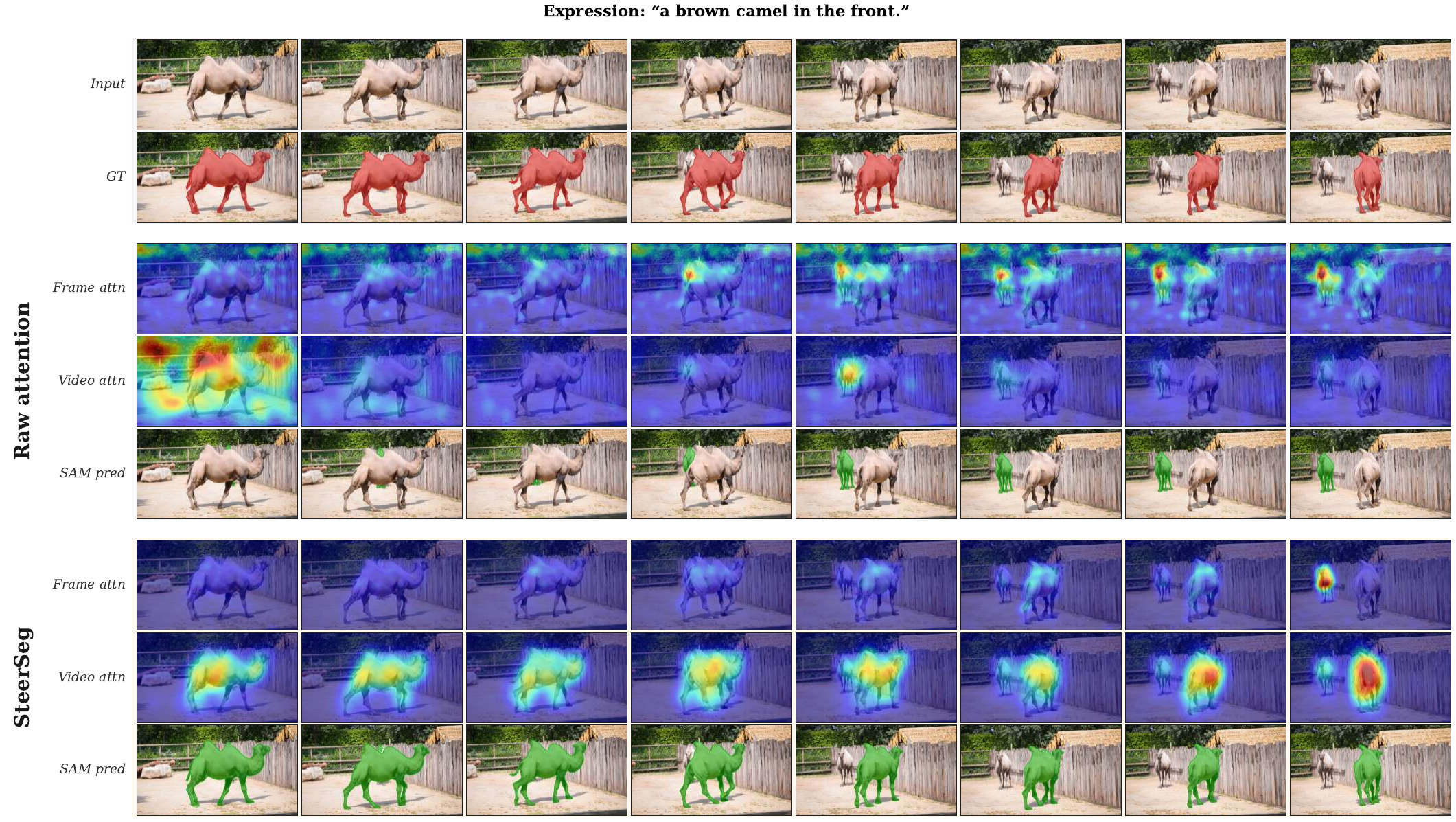}
\end{figure}

\begin{figure}[htbp]
    \centering
    \includegraphics[width=\linewidth]{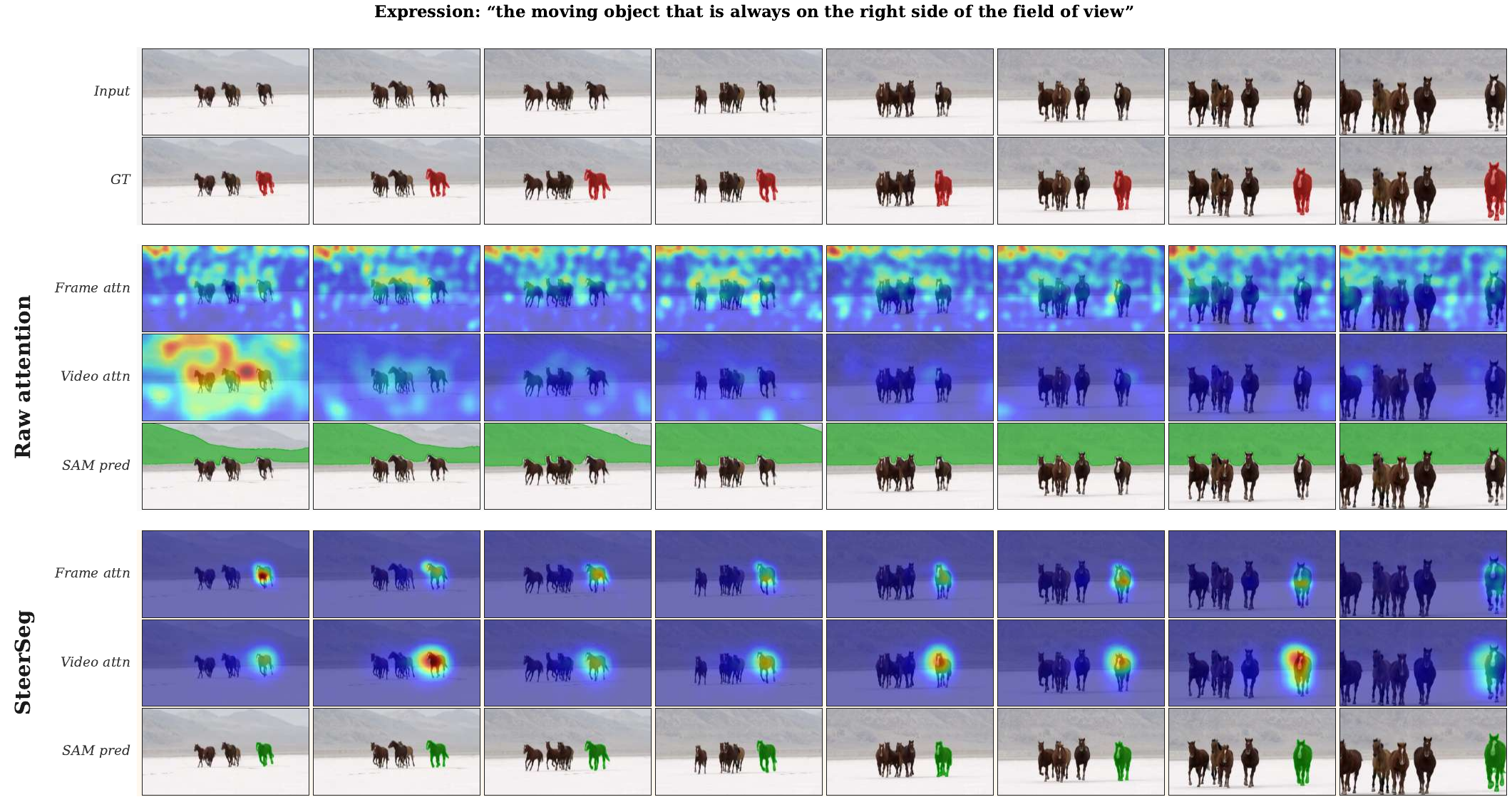}
\end{figure}


\begin{figure}[htbp]
    \centering
    \includegraphics[width=\linewidth]{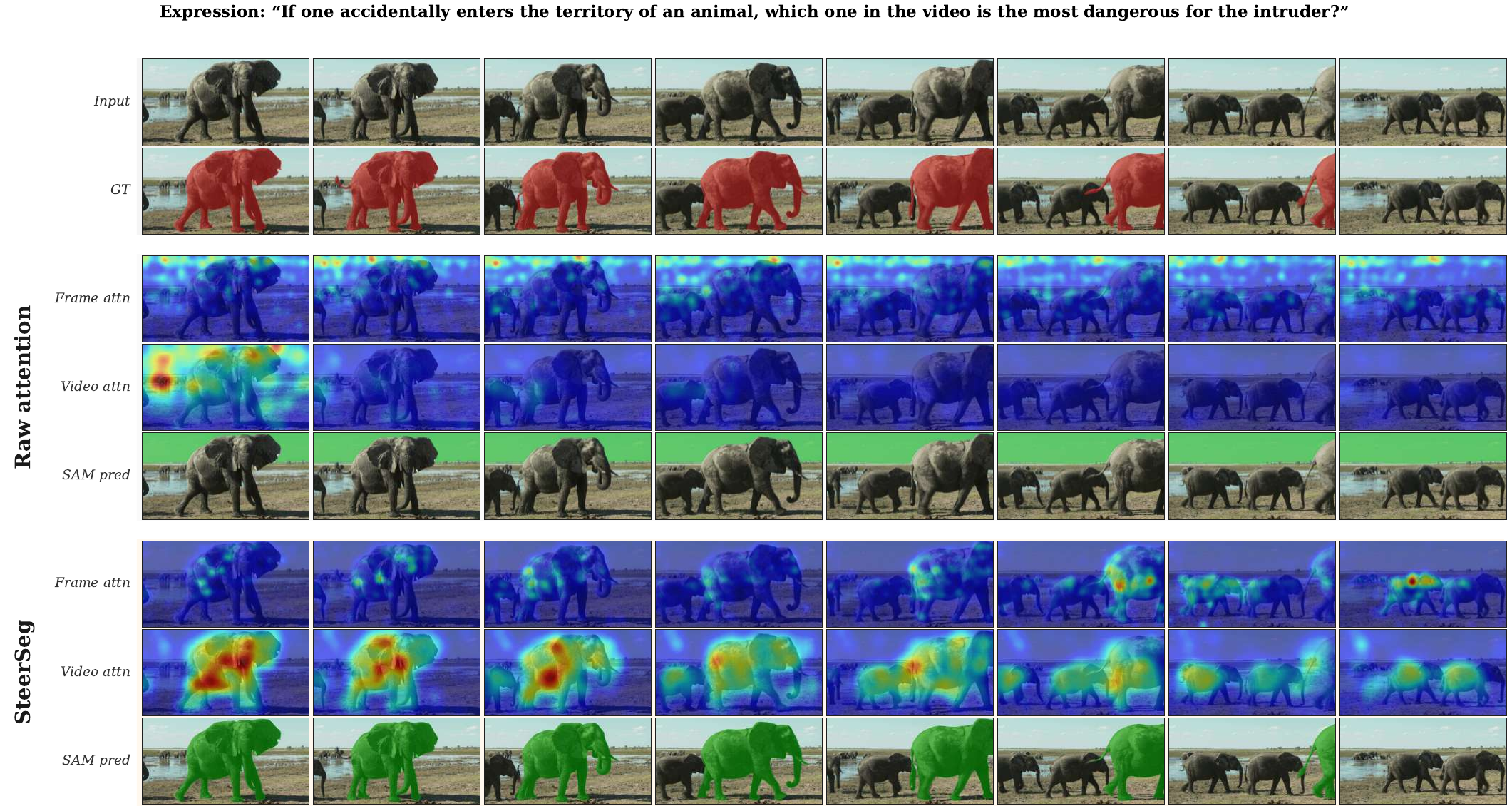}
\end{figure}

\begin{figure}[htbp]
    \centering
    \includegraphics[width=\linewidth]{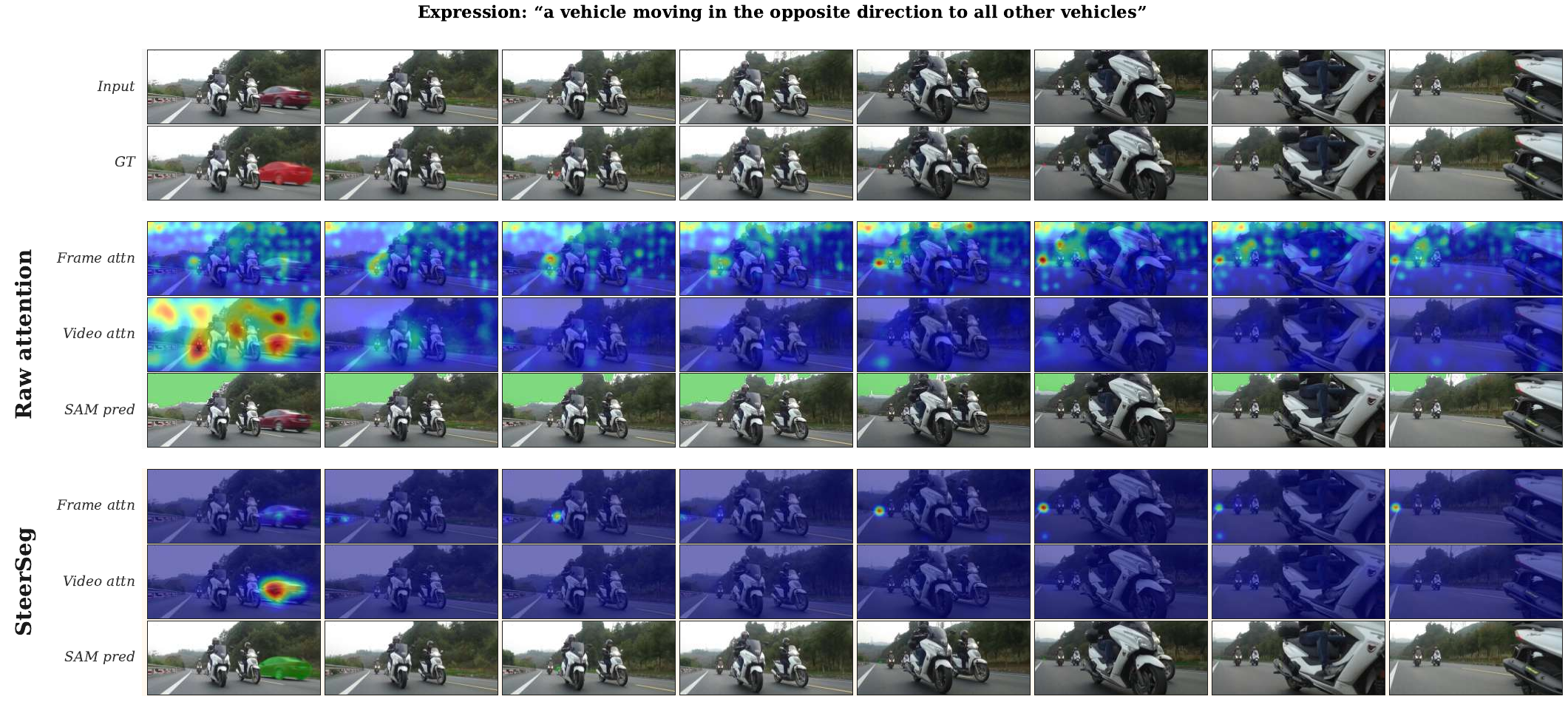}
\end{figure}




\clearpage
\newpage
\newpage

\end{document}